\documentclass[10pt,twocolumn,letterpaper]{article}

\usepackage{iccv}              

\definecolor{iccvblue}{rgb}{0.21,0.49,0.74}
\usepackage[pagebackref,breaklinks,colorlinks,allcolors=iccvblue]{hyperref}

\usepackage{listings}
\usepackage{cuted}
\usepackage{multirow}
\usepackage{pifont}
\usepackage{dsfont}
\usepackage{xcolor}         
\usepackage{graphicx}

\def\paperID{6865} 
\def\confName{ICCV}
\def\confYear{2025}

\title{Controllable 3D Outdoor Scene Generation via Scene Graphs}

\newcommand*{\affaddr}[1]{#1} 
\newcommand*{\affmark}[1][*]{\textsuperscript{#1}}

\makeatletter
\renewcommand{\@makefntext}[1]{\noindent#1}
\makeatother

\author{%
Yuheng Liu\affmark[1,2], 
Xinke Li\affmark[3]$^\dag$, 
Yuning Zhang\affmark[4], 
Lu Qi\affmark[5],
Xin Li\affmark[1],
Wenping Wang\affmark[1], \\
Chongshou Li\affmark[4], 
Xueting Li\affmark[6]$^*$, 
Ming-Hsuan Yang\affmark[2]$^*$
\vspace{2mm}
\\
\affaddr{\affmark[1]Texas A\&M University}, 
\affaddr{\affmark[2]UC Merced}, 
\affaddr{\affmark[3]City University of Hong Kong} \\
\affaddr{\affmark[4]Southwest Jiaotong University}, 
\affaddr{\affmark[5]Insta360 Research}, 
\affaddr{\affmark[6]NVIDIA} 
\vspace{-3.5em}
}

\begin{document}
\maketitle

\footnotetext[1]{$^\dag$ Corresponding author.}
\footnotetext[2]{$^*$ Equal contribution.}

\begin{strip}
  \centering
\vspace{-1em}
\includegraphics[width=0.80\linewidth]{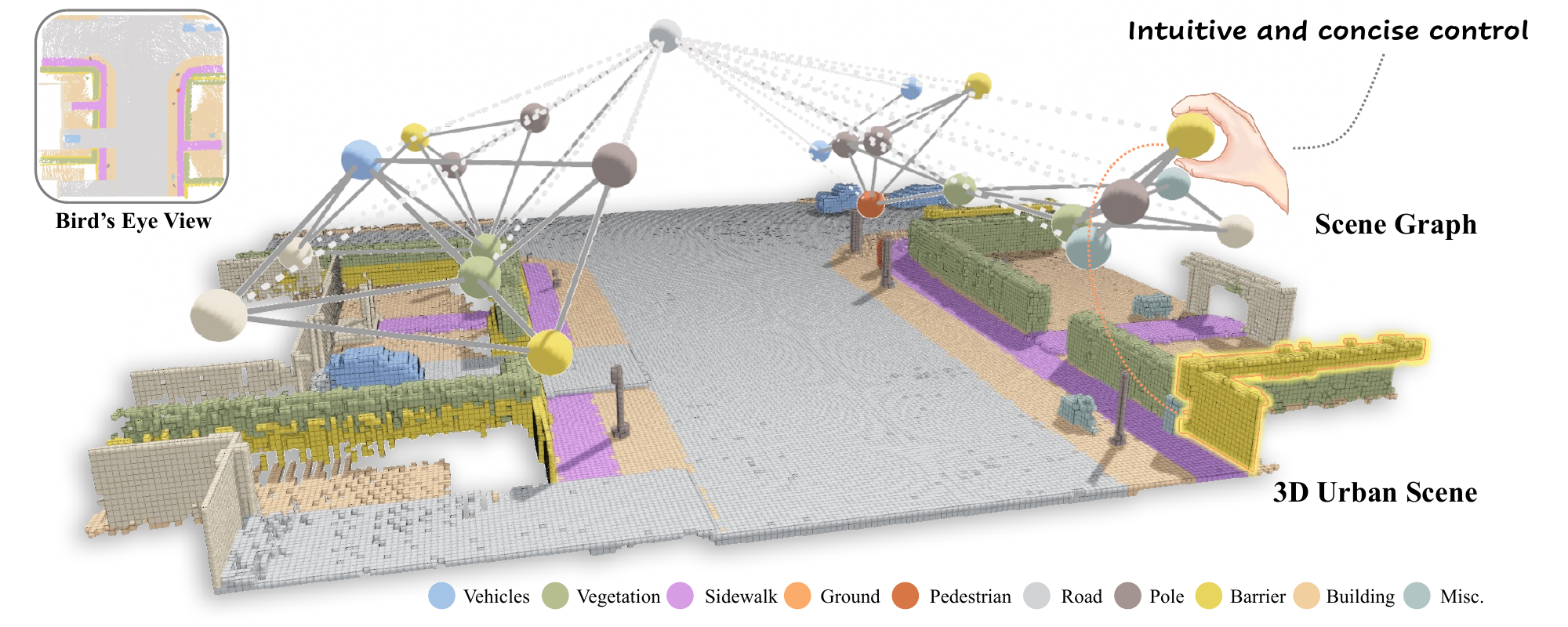}
\captionof{figure}{\textbf{Scene Graph Guided 3D Outdoor Scene Generation.} Compared to text descriptions and BEV maps, scene graphs offer a more intuitive and user-friendly format for controlling 3D scene generation. We also develop an interactive system that allows users to generate/edit dense 3D scenes through scene graph interaction.}
\vspace{-6pt}
\label{fig:teaser}
\end{strip}

\begin{abstract}
Three-dimensional scene generation is crucial in computer vision, with applications spanning autonomous driving, gaming and the metaverse. Current methods either lack user control or rely on imprecise, non-intuitive conditions. In this work, we propose a method that uses scene graphs—an accessible, user-friendly control format—to generate outdoor 3D scenes. We develop an interactive system that transforms a sparse scene graph into a dense BEV (Bird's Eye View) Embedding Map, which guides a conditional diffusion model to generate 3D scenes that match the scene graph description. During inference, users can easily create or modify scene graphs to generate large-scale outdoor scenes. We create a large-scale dataset with paired scene graphs and 3D semantic scenes to train the BEV embedding and diffusion models. Experimental results show that our approach consistently produces high-quality 3D urban scenes closely aligned with the input scene graphs. To the best of our knowledge, this is the first approach to generate 3D outdoor scenes conditioned on scene graphs. 
\end{abstract}

\section{Introduction}
\label{sec:introduction}
3D scene generation has garnered wide attention due to its potential for creating realistic, physically coherent 3D scenes. These models offer a powerful approach to understanding and simulating the complexities of our 3D world. Among the various methods for 3D scene generation, probabilistic generative models have shown great promise in recent advancements. However, the stochastic nature of these models makes the generation process difficult to control precisely, emphasizing the need for an editable and controllable generation process.

To enable controllable scene generation,
many methods leverage recent advances in 2D conditional generation, such as DALL-E~\cite{betker2023improving} and Stable Diffusion~\cite{rombach2022high}, where high-quality images are generated based on natural language. Inspired by these models, some approaches~\cite{lin2023magic3d, liu2024isotropic3d, liu2024one} use 2D views to guide 3D content generation. However, these methods are primarily object-centric and do not scale to complex outdoor scenes due to their large scales and interconnected structures. Other approaches apply text-based conditions to directly control 3D scene generation, such as Text2LiDAR~\cite{wu2024text2lidar}. Yet, text-to-3D provides insufficient control over both physical constraints and spatial details: it cannot effectively enforce real-world physical rules or precisely control scene elements (e.g., number of objects), leading to outputs that fail to meet specified requirements~\cite{zhang2023adding}.

One possible solution is to extend existing 3D indoor scene generation methods~\cite{zhai2025echoscene, zhai2023commonscenes, wei2024planner3d, yang2025mmgdreamer, hu2024mixed} to outdoor environments.
However, this adaptation is highly challenging. Indoor scene generation typically relies on multi-view images to synthesize bounded, textured surfaces, focusing on object appearance and spatial relationships. In contrast, outdoor scenes are unbounded and predominantly captured using textureless LiDAR point clouds, 
which aims to model large-scale spatial layouts and background continuity.

Recent outdoor scene generation research attempts to explore unique controls for 3D outdoor scene generation. For example, 
~\cite{zhang2024urban, deng2023citygen} rely on BEV layouts or semantic maps, which require users to provide pixel-level control signals, posing a challenge in terms of interaction, especially for large-scale, complex 3D outdoor scenes. Therefore, selecting an appropriate medium for controllable 3D outdoor scene generation is crucial. In this context, the scene graph emerges as an ideal candidate due to its structured, regularized, and sparse representation, which makes it particularly well-suited for 3D outdoor scene generation and allows efficient control over complex layouts.
Additionally, scene graphs are intuitive, enabling users to interact with and edit them easily. Motivated by these benefits, we propose a new framework that leverages the scene graph as a sparse-to-dense pipeline for 3D outdoor scene generation.

However, it is non-trivial to utilize scene graph as condition, due to its sparse and abstract nature. To resolve this, 
we begin by employing a Graph Neural Network (GNN) that aggregates information from the scene graph through message passing. Next, a novel Allocation Module is developed to assign spatial positions to produce a Bird's Eye View Embedding Map (BEM). Finally, the BEM is used to condition a 3D Pyramid Discrete Diffusion Model~\cite{pdd} to generate the complete 3D scene. 
We jointly train the GNN and the diffusion model for seamless integration. 
To enhance scene understanding within the GNN, two auxiliary tasks are introduced: edge reconstruction and node classification, which further improve the model's ability to interpret and represent the scene graph effectively. 
Additionally, we develop an interactive system to enable intuitive scene graph creation and editing, allowing users to control scene content through both manual editing and text-based scene graph generation, bridging the gap between text input and 3D outdoor scene generation. To support our approach, we construct a scene graph dataset for each 3D scene in the CarlaSC dataset~\cite{carlasc}, defining node attributes and establishing edges based on spatial relationships. The primary contributions of our work are as follows:
\begin{itemize}
\item To the best of our knowledge, this is the first work that generates a large-scale 3D outdoor scene conditioning on a scene graph input.
\item We propose a GNN equipped with a novel Allocation Module that converts a sparse scene graph to a compact scene embedding, which then conditions a diffusion model for 3D scene generation.
\item We curate a large-scale dataset including paired 3D scenes and scene-graphs for model training. Extensive experiments demonstrate that our approach generates 3D outdoor scenes that closely align with the definitions in the scene graphs.
\item We develop and provide a user-friendly system for constructing scene graphs, enabling users to flexibly create custom scene graphs to guide 3D outdoor scene generation according to their needs.
\end{itemize}

\begin{figure*}[t!]
  \centering
  \includegraphics[width = 1.0\textwidth]
  {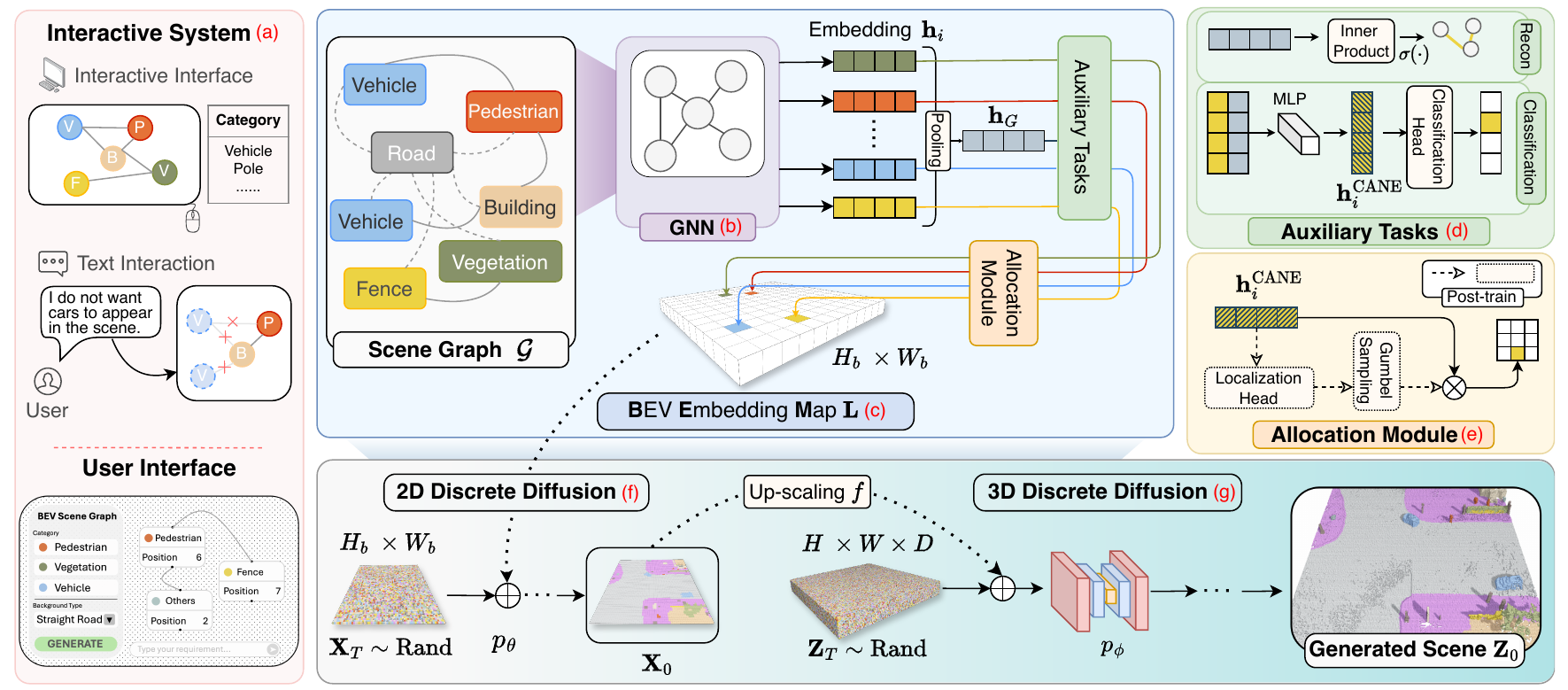}
  \vspace{-2em}
    \caption{\textbf{Overview of Scene Graph Guided 3D Scene Generation.} The Scene Graph Guided 3D Generation structure consists of three main components: the interactive system (red), BEM processing (blue), and diffusion generation (bottom). Through the interactive system, users can construct their own Scene Graphs using either an interactive interface or text interaction. The constructed scene graph is processed by a GNN, which is jointly trained with the diffusion model using auxiliary tasks to enhance control. Each node in the Scene Graph is then positioned by the Allocation Module to form the BEM. This BEM serves as a conditioning input to the 3D Pyramid Discrete Diffusion Model~\cite{pdd}, which generates the final 3D outdoor scene. Note that ``Recon", ``Classification", and ``CANE" denote ``Edge Reconstruction", ``Node Classification", and ``Context-aware Node Embedding", respectively.}
    \label{fig:main_method}
    \vspace{1em}
\end{figure*}

\section{Related Work}
\label{sec:related_work}

\noindent \textbf{3D Generation via Diffusion Models.} 
Diffusion models have expanded their applications from 2D image synthesis to complex 3D data modeling~\cite{pamiDMiV2023}. 
Compared to conventional GANs~\cite{GAN} and VAEs~\cite{VAE}, they offer improved performance through a progressive denoising mechanism~\cite{DDPM}, enhancing training stability and the ability to model complex distributions.
This makes them particularly effective for 3D data generation. 
However, existing research primarily focuses on object-level generation~\cite{xu2025sparp, ren2024xcube, magic123, liu2023zero1to3, yang2019pointflow, xiang2024structured} or indoor environments~\cite{Bokhovkin2024SceneFactorFL, tang2024diffuscene, Yang2024SceneCraftL3, Fang2023CtrlRoomCT}. 
Meanwhile, the few works~\cite{lin2023infinicity, xie2024citydreamer,pdd,lee2024semcity,lee2023diffusion} designed for outdoor scene generation prioritize visual fidelity over controllability. 
In this work, we aim to develop a framework for controllable 3D outdoor scene generation, emphasizing easy and precise user control.

\noindent \textbf{Scene Graph Application.} 
A scene graph is a structured representation of a scene, encoding objects, their attributes, and the relationships between them~\cite{Johnson_2015_CVPR}. 
Unlike dense representations such as point clouds~\cite{qi2017pointnet++,li2025deephierarchicallearningfor3dsemanticsegmentation,li2024enhancing} or meshes~\cite{chang2015shapenet}, a scene graph provides a concise yet comprehensive view of a scene’s structure, making it a powerful conditional signal for generation tasks. 
By explicitly modeling inter-object relationships, scene graphs offer a structured way to control scenes, facilitating both human-driven and AI-driven content generation, and this capability has been widely explored especially in 2D tasks~\cite{SonSGpami2023, Liu2024R3CDSG, yang2022dbsgi, Yang_2019_CVPR, XU2019477, vcg2021tmm, johnson2018image, farshad2023scenegenie}. 
Extending this concept to 3D environments, Armeni et al.~\cite{armeni_iccv19} introduce the 3D indoor scene graph. This framework integrates semantic, spatial, and geometric information into a hierarchical graph, where nodes represent objects, rooms, and spaces, and edges encode their spatial and semantic relationships. Although this well-defined representation has proven effective for indoor scenes, a comparable formulation for outdoor environments remains largely unexplored.

\noindent \textbf{Controllable 3D Scene Generation.} 
To date, controllable 3D outdoor scene generation has received limited attention. One of the closest efforts in this area is Text2LiDAR~\cite{wu2024text2lidar}, which generates LiDAR points from text inputs.
While text-based control is an interesting attempt, it lacks explicit spatial structure, making it unsuitable for precise scene composition~\cite{zhang2023adding}.
Instead, scene graphs offer a more interpretable and structured approach to controlling scene generation, a concept that has been well established in indoor scene generation~\cite{lin2024instructscene, tang2024diffuscene, zhou2019scenegraphnet, li2019grains, wang2021sceneformer}. 
However, adapting them to outdoor scenes is non-trivial, as outdoor scenes are large-scale, unbounded, and contain a diverse set of objects~\cite{xie2025citydreamer4dcompositionalgenerativemodel}. 
Unlike indoor scenes, which often use a compositional approach by placing objects within predefined or generated bounding boxes~\cite{zhai2025echoscene, zhai2023commonscenes, dhamo2021graph, wang2019planit, para2021generative}, outdoor scenes feature complex backgrounds and incomplete structures, such as roads and buildings.
To address these challenges, we propose a novel pipeline and scene graph representation tailored for controllable 3D outdoor scene generation.

\section{Method}
\label{sec:approach}

We first discuss the formulation of a scene graph in Sec.~\ref{sec:sg_formulation}. We then introduce how to generate a 3D outdoor scene conditioned on a scene graph in Sec.~\ref{sec:scene-graph-guided}. In Sec.~\ref{sec:interactive}, we describe how the proposed method facilitates the convenient creation of 3D scenes.

\subsection{Scene Graph Formulation}
\label{sec:sg_formulation}
Formally, a scene graph is characterized by its nodes and edges as \( \mathcal{G} = (\mathcal{V}, \mathcal{E}) \).
The set of nodes \( \mathcal{V} \) consists of two types (i.e., \( \mathcal{V} = \mathcal{V}_I \cup \mathcal{V}_R \)):
a) \textbf{Instance Nodes} (\( \mathcal{V}_I \)) that represent countable objects with standard labels defined in ~\cite{carlasc}, such as vehicles and pedestrians. 
Each node \( v_i \in \mathcal{V}_I \) is associated with a feature vector $[\mathbf{c}_i; \mathbf{p}_i]$, where \( \mathbf{c}_i \in \mathbb{R}^d \) denotes the node attributes with dimension $d$, and \( \mathbf{p}_i \in \mathbb{R}^2 \) represents a coordinate specifying its center position in the BEV map. 
b) \textbf{Scene Road Nodes} (\( \mathcal{V}_R \)) that define the structure of the road and other global background information of the scene in one node, i.e., $\mathcal{V}_R=\{v_r\}$. 
We further construct the graph structure by defining two types of edges \( \mathcal{E} \) to capture essential relationships: a) \textbf{Physical Proximity}: For any two instance nodes \( v_i, v_j \in \mathcal{V}_I \), we define an edge \( e_{ij} \in \mathcal{E} \) if the Euclidean distance \( d_{ij} = \|\mathbf{p}_i - \mathbf{p}_j\| \)  is smaller than a threshold \( \delta_d \). b) \textbf{Road Connectivity}: For any instance node \( v_i \in \mathcal{V}_I \) and the singleton road node \( v_r \in \mathcal{V}_R \), an edge \( e_{ir} \in \mathcal{E} \) is created for the connects to the road structure. 

In practice, we use the simplified graph as control signals to ease user interaction, where each instance node contains only its semantic label $\mathbf{c}_i$ and an approximate 2D position $\mathbf{p}_i$ represented as a patch index. Also, the scene road node is represented by the road type.

\subsection{Scene-graph-guided Diffusion}
\label{sec:scene-graph-guided}

\begin{figure}[t]
    \centering
    \includegraphics[width = 0.97\linewidth]
    {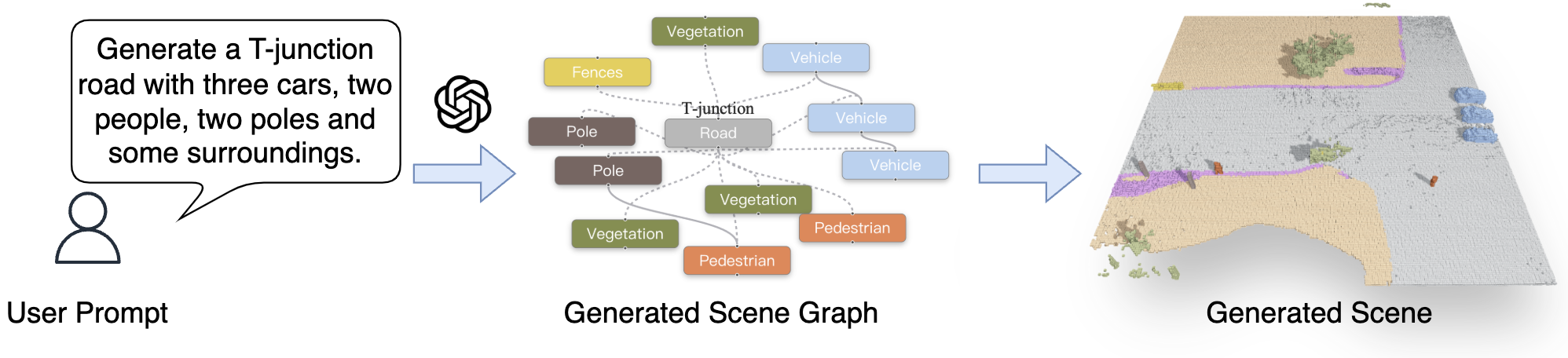}
    \vspace{-0.5em}
    \caption{\textbf{Scene Graph Generation.} LLMs convert the user’s prompt into a scene graph, which guides 3D scene generation.}
    \label{fig:prompt_generation}
    \vspace{-1em}
\end{figure}

Given a 2D scene graph, our method aims at generating a 3D semantic scene that aligns with the structure defined by the scene graph. In the following, we first convert the scene graph into a dense 2D embedding using a Graph Neural Network (GNN). Next, we synthesize a plausible 2D scene map by training a 2D diffusion model conditioned on the scene graph embedding. Finally, we use a conditional 3D diffusion model to generate the final 3D outdoor scene given the 2D scene map. 

\noindent \textbf{Scene Graph Neural Network}. 
\label{sec:gnn}
The scene Graph Neural Network (Fig.~\ref{fig:main_method}\textcolor{red}{(b)}) aims to generate node embeddings for scene graphs that capture both local structural and global context information. Particularly, we utilize  Graph Attention Network (GAT) \cite{gat} for GNN implementation.
Given a graph \( \mathcal{G} = (\mathcal{V}, \mathcal{E}) \),  with adjacency matrix \( \mathbf{A} \in \mathbb{R}^{|\mathcal{V}| \times |\mathcal{V}|} \) that encodes the connectivity between nodes, 
the node embeddings $\mathbf{h}_i$ are computed by a two-layer GAT for node \( v_i \) in the graph.  
To incorporate global context into the node embeddings, we concatenate each node’s embedding with a pooled global embedding \( \mathbf{h}_G \) as the final embedding, i.e.,
\small
\begin{equation}
   \mathbf{h}_i^{\text{CANE}} = \text{MLP}([\mathbf{h}_i;\mathbf{h}_G 
   ]), \mathbf{h}_G= \text{Pooling}(\{\mathbf{h}_i \mid v_i \in \mathcal{V}\}),
\end{equation}
\normalsize
where  \( [\cdot; \cdot] \) denotes concatenation,  \( \text{Pooling} (\cdot)\) is a graph global mean pooling operation~\cite{mesquita2020rethinking}, $\mathbf{h}_G \in \mathbb{R}^{64}$ and $\text{MLP}(\cdot)$ is a multi-layer perceptron. We name the output embedding $\mathbf{h}_i^{\text{CANE}}$ as Context-aware node embedding (CANE).

We train the GNN with two objectives: the auxiliary tasks (Fig.~\ref{fig:main_method}\textcolor{red}{(d)}) and the downstream task (Fig.~\ref{fig:main_method}\textcolor{red}{(e)}). 
For auxiliary tasks, we apply edge reconstruction loss as in Graph Auto-encoder (GAE)~\cite{gae} and node classification loss, given by
\small
\begin{equation}
\label{eq:a_loss}
     \mathcal{L}_a = \text{BCE}(\hat{\mathbf{A}}, \mathbf{A})+  \frac{1}{|\mathcal{V}|} \sum_{i=1}^{|\mathcal{V}|} \text{CE} (y_i, \hat{y}_i),
\end{equation}
\normalsize
where $\text{BCE}$ is binary cross-entropy, $\text{CE}$ is node-wise  cross-entropy loss, and,
\small
\begin{equation}
    \hat{\mathbf{A}} = \sigma(\mathbf{h}_G \mathbf{h}_G^\top), \hat{y}_i = \text{Softmax}(\text{MLP} (\mathbf{h}_i^\text{CANE})),
\end{equation} 
\normalsize
for $\sigma(\cdot)$ being sigmoid function.
The first term of ~\eqref{eq:a_loss} is to reconstruct the global edge structure of the scene, specifically the adjacency matrix \( \mathbf{A} \). The second term of ~\eqref{eq:a_loss} is to classify each node \( v_i \) into the original category. 
The auxiliary tasks ensure that the network learns both structural relationships and node-specific features, effectively capturing both local and global information in the CANE. 

For downstream task,  embeddings \( \mathbf{h}_i^\text{CANE} \) are used as inputs to compute BEV Embedding Map (BEM, Fig.~\ref{fig:main_method}\textcolor{red}{(c)}) which serves as condition for the sequential diffusion model in the next phase of diffusion. 
These embeddings are passed to an \textbf{allocation module},  defined by
\small
\begin{equation}\label{eq:allocation}
    \mathbf{L} = \sum_{i=1}^{|\mathcal{V}|} \mathcal{M}(\hat{p}_i) \odot \mathbf{h}_i^\text{CANE},
\end{equation}
\normalsize
where $\mathbf{L}\in \mathbb{R}^{H_b\times W_b\times C} $ is the output BEM with height $H_b$, width $W_b$, and channel dimension $C$. The binary map $ \mathcal{M}(\hat{p}_i)\in \{0,1\}^{H_b\times W_b}$ is expanded along the channel dimension to \( \mathbb{R}^{H_b \times W_b \times C} \) to enable element-wise multiplication with the node embedding \( \mathbf{h}_i^\text{CANE} \in \mathbb{R}^{C} \).  
In the implementation, we perform the inference by sampling position from an MLP-based localization head, i.e.,
\small
\begin{equation}
      \hat{p}_i \sim \text{GumbelSoftmax}_{\tau}(\text{Head}(\mathbf{h}_i^\text{CANE})),
\end{equation}
\normalsize
where $\tau$ is the temperature for Gumbel softmax~\cite{jang2016categorical}. While in the training process of the diffusion model, we replace the $\hat{p}_i$ in ~\eqref{eq:allocation} by the ground truth position $p_i$. The localization head is trained after the diffusion model training. 
As a result, the allocation module effectively converts irregular and sparse graph representation into dense 2D map, i.e., BEM, which offers better compatibility with the subsequent 2D diffusion process.

\begin{figure*}[!t]
  \centering
  \includegraphics[width = 0.99\textwidth]
  {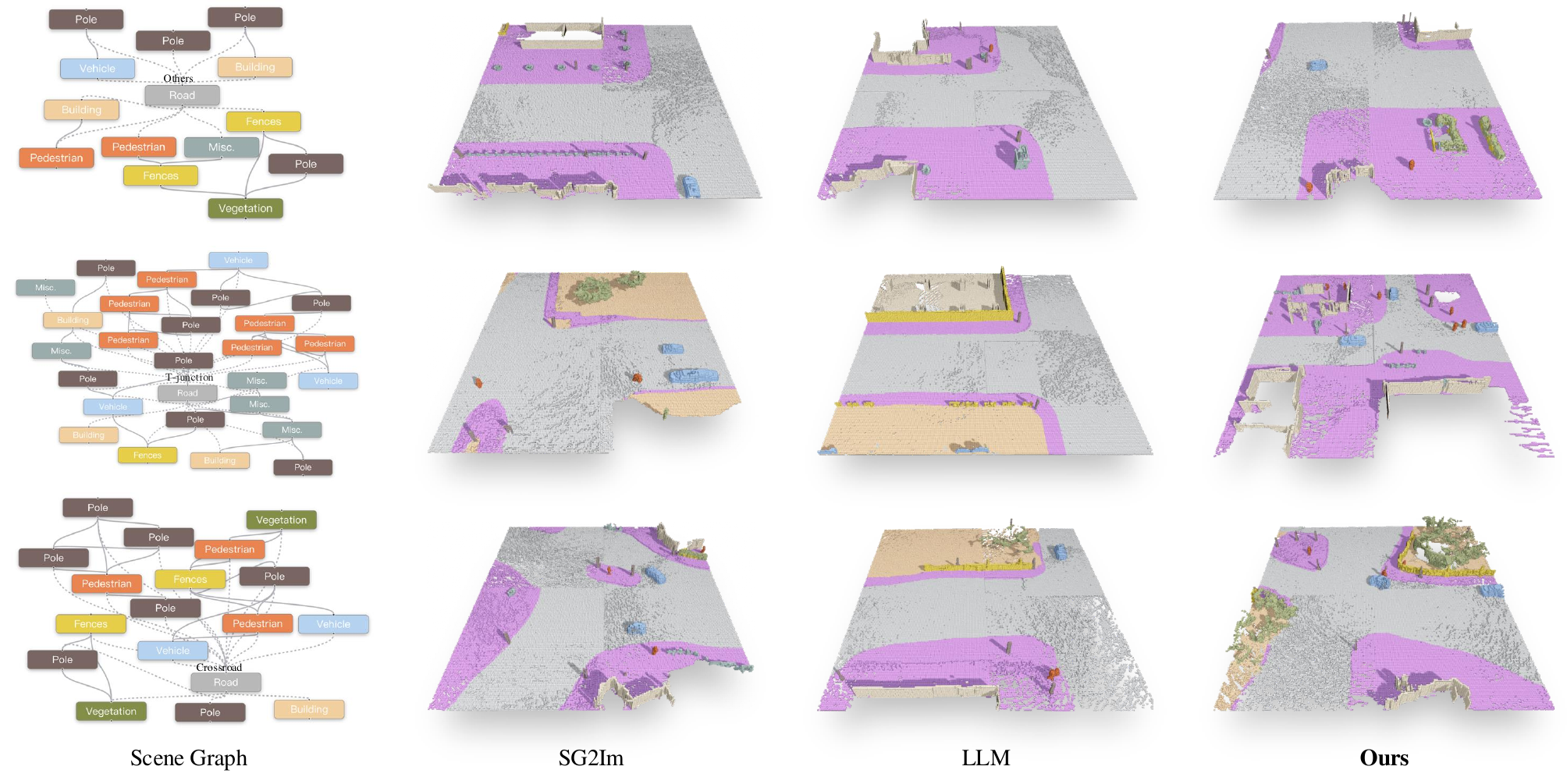}
  \vspace{-1em}
    \caption{\textbf{Controlling 3D Outdoor Scene Generation with Scene Graphs.} We compare baseline methods. Results show that our method generates scenes consistent with the provided scene graph, whereas the SG2Im and LLM approaches exhibit inconsistencies in object quantities and road types. 
    }
    \label{fig:main_comparison}
    \vspace{0em}
\end{figure*}

\noindent \textbf{2D Map Discrete Diffusion (Fig.~\ref{fig:main_method}\textcolor{red}{(f)}).} Given a scene graph, there may be multiple plausible 2D maps that align with its structure. To model this variability, we use a diffusion model that converts the sparse scene graph embedding into a dense 2D map representation.
Formally, the 2D Map Diffusion refines the sparse BEM $\mathbf{L}$ into a dense 2D map,  $\mathbf{X} \in \{0, 1\}^{H_b \times W_b \times c}$, where \( c \) is the semantic class number. We apply the standard discrete diffusion~\cite{austin2021structured} for 2D map generation.  
In the forward process, the 2D map \( \mathbf{X}_0 \) is gradually corrupted in \( T \) timesteps using a transition matrix \( \mathbf{Q}_t \), which adds noise as $\mathbf{X}_t = \mathbf{X}_{t-1} \mathbf{Q}_t$. 
This process can also be represented using a cumulative matrix \( \bar{\mathbf{Q}}_t \), allowing us to sample \( \mathbf{X}_t \) directly from the original map \( \mathbf{X}_0 \), 
\small
\begin{equation}
   q(\mathbf{X}_t \mid \mathbf{X}_0) = \operatorname{Cat}(\mathbf{X}_t; \mathbf{P} = \mathbf{X}_0 \bar{\mathbf{Q}}_t),
\end{equation}
\normalsize
where \( \operatorname{Cat} \) represents a categorical distribution.

In the reverse diffusion stage, a model \( p_\theta \) learns to reverse the noise process, predicting the less-noised map \( \mathbf{X}_{t-1} \) from the noisy map \( \mathbf{X}_t \), using \( \mathbf{L} \) as guidance:
\small
\begin{equation}
   p_\theta(\mathbf{X}_{t-1} \mid \mathbf{X}_t, \mathbf{L}) = \mathbb{E}_{\tilde{p}_\theta(\tilde{\mathbf{X}}_0 \mid \mathbf{X}_t, \mathbf{L})} q(\mathbf{X}_{t-1} \mid \mathbf{X}_t, \tilde{\mathbf{X}}_0).
\end{equation}
\normalsize
The model is trained by minimizing the KL divergence between the forward process and the learned reverse process. The loss function \( \mathcal{L}_\theta \) is defined as:
\small
\begin{align}\label{eq:diff_loss}
   \mathcal{L}_\theta =   & d_{\text{KL}}\left(q(\mathbf{X}_{t-1} \mid \mathbf{X}_t, \mathbf{X}_0) \| p_\theta(\mathbf{X}_{t-1} \mid \mathbf{X}_t, \mathbf{L})\right) \\\notag
   &+ \lambda d_{\text{KL}}\left(q(\mathbf{X}_0) \| \tilde{p}_\theta(\tilde{\mathbf{X}}_0 \mid \mathbf{X}_t, \mathbf{L})\right),
\end{align}
\normalsize
where \( \lambda \) controls the weight of the auxiliary term for better reconstruction. During inference, we start from random noise and use the learned reverse diffusion to generate a dense 2D map \( \mathbf{X}_0 \), guided by the BEM \( \mathbf{L} \). This refined map provides a more complete spatial layout for further 3D scene generation. We train the GNN and the 2D map diffusion model using a loss 
$\mathcal{L}_a+\mathcal{L}_{\theta}$. 

\noindent \textbf{3D Scene Discrete Diffusion (Fig.~\ref{fig:main_method}\textcolor{red}{(g)}).} We convert the generated 2D map into a dense 3D scene using a discrete diffusion process similar to the 2D Map Diffusion. The initial 2D map \( \mathbf{X}_0 \in \{0, 1\}^{H_d \times W_d \times c} \), generated from the previous diffusion step, is used as a condition to guide the generation of the 3D scene. We define the 3D scene as \( \mathbf{Z} \in \{0, 1\}^{H \times W \times D \times c} \), where \( H \), \( W \), and \( D \) are the dimensions of the 3D scenes and \( c \) represents semantic categories.

The 3D diffusion follows the same forward and reverse diffusion steps as in the 2D case but operates on the 3D scene grid. Specifically, a learnable model \( p_\phi \) predicts each denoised state \( \mathbf{Z}_{t-1} \) from \( \mathbf{Z}_t \), conditioned on the input 2D map \( \mathbf{X}_0 \), given by,
\small
\begin{equation}\label{eq:prob_3}
     p_\phi(\mathbf{Z}_{t-1} \mid \mathbf{Z}_t, \mathbf{X}_0)=\mathbb{E}_{\tilde{p}_\theta(\tilde{\mathbf{Z}}_0 \mid \mathbf{Z}_t, f(\mathbf{X}_0))} q(\mathbf{Z}_{t-1} \mid \mathbf{Z}_t, \tilde{\mathbf{Z}}_0),
\end{equation}
\normalsize
where $f:\mathbb{R}^{H_d\times\ W_d\times c}\rightarrow \mathbb{R}^{H\times\ W \times c}$ is an up-scaling function. Furthermore, the training loss \( \mathcal{L}_\phi \) follows the same form as \eqref{eq:diff_loss}, given by,
\small
\begin{align}\label{eq:diff_loss2}
   \mathcal{L}_\phi =   & d_{\text{KL}}\left(q(\mathbf{Z}_{t-1} \mid \mathbf{Z}_t, \mathbf{Z}_0) \| p_\phi(\mathbf{Z}_{t-1} \mid \mathbf{Z}_t, \mathbf{X}_{0})\right) \\\notag
   &+ \lambda d_{\text{KL}}\left(q(\mathbf{Z}_0) \| \tilde{p}_\phi(\tilde{\mathbf{Z}}_0 \mid \mathbf{Z}_t, \mathbf{X}_{0})\right).
\end{align}
\normalsize
During inference, the network generates the final 3D scene \( \mathbf{Z}_0 \) by starting from a noisy 3D state and applying the reverse diffusion process conditioned on \( \mathbf{X}_0 \). Specifically, each step of the reverse diffusion process is performed by sampling as ~\eqref{eq:prob_3}.
This produces a fully detailed 3D scene aligned with the spatial layout of the 2D map.

\begin{table*}[!t]
\centering
\caption{\textbf{Comparison of Different Conditioning Methods on 3D Outdoor Scene Generation.} Uncon-Gen, SG2Im, and LLM represent Unconditional Generation, Scene Graph to Image, and Large Language Model, while M-Pole, M-Pede, and M-Vech represent the MAE calculated individually for \textit{Pole}, \textit{Pedestrian}, and \textit{Vehicle} categories. In the Scene Quality Evaluation, higher mIoU and MA scores indicate better semantic consistency, while a lower F3D score~\cite{pdd} signifies closer feature alignment with the original dataset. In the Control Capacity Evaluation, a lower MAE reflects a smaller discrepancy between the generated scene and the object quantities defined in the scene graph for conditioning. A higher Jaccard Index indicates greater alignment in the object categories between the generated scenes and the specified scene graph.}
\vspace{-0.5em}
\resizebox{1\textwidth}{!}{
\begin{tabular}{cc|ccc|ccccc}
\toprule
\multirow{2}{*}{\textbf{Method}} & \multirow{2}{*}{\textbf{Condition}} & \multicolumn{3}{c|}{\textbf{Scene Quality}}               & \multicolumn{5}{c}{\textbf{Control Capacity}}                                                                     \\
                        &                            & mIoU           & MA             & \multicolumn{1}{c|}{F3D (\(\downarrow\))}               & MAE (\(\downarrow\)) & Jaccard       & M-Pole (\(\downarrow\)) & M-Pede (\(\downarrow\)) & M-Vech (\(\downarrow\)) \\ \midrule \midrule
Uncon-Gen~\cite{pdd}               & -                          & 68.21          & \textbf{85.69} & \multicolumn{1}{c|}{\textbf{0.338}}          & 2.07                 & 0.68          & 4.78                    & 4.71                    & 3.59                    \\ \midrule
SG2Im~\cite{johnson2018image}               & Scene Graph              & 65.43          & 81.72          & \multicolumn{1}{c|}{0.486} & 0.97                 & 0.81          & 2.25                    & 2.79                    & 2.64                    \\
LLM~\cite{ye2024language, zhao2023graphtext}               & Text-Embedding              & 68.19          & 85.62          & \multicolumn{1}{c|}{0.386} & 1.44                 & 0.70          & 3.41                    & 3.57                    & 3.51                    \\
\textbf{Ours}              & Scene Graph            & \textbf{68.69} & 85.01          & \multicolumn{1}{c|}{0.393}          & \textbf{0.63}        & \textbf{0.93} & \textbf{1.39}           & \textbf{1.81}           & \textbf{1.35}           \\ \bottomrule
\end{tabular}}
\label{tab:main_results}
 \vspace{0em}
\end{table*}

\begin{figure*}[t]
  \centering
  \includegraphics[width = 1\textwidth]
  {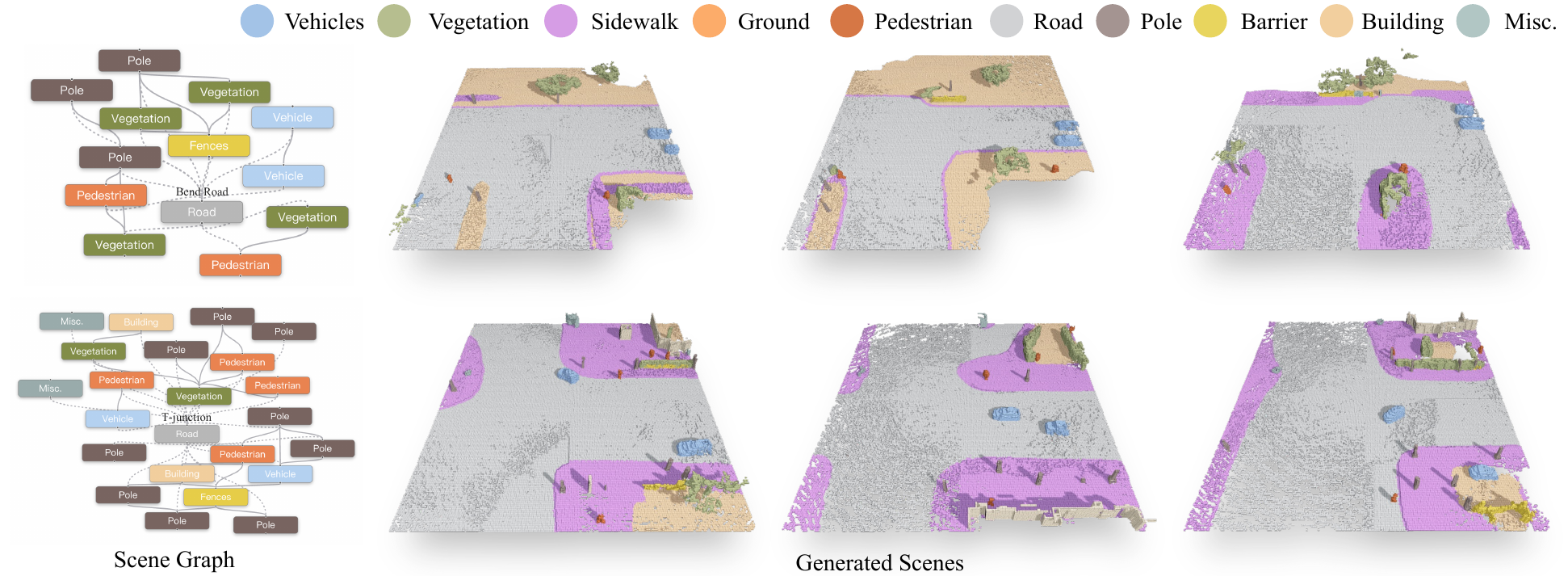}
  \vspace{-2em}
    \caption{\textbf{Diversity in Scene Generation.} Comparison of three scenes generated by our method under the same scene graph. This demonstrates our method’s ability to produce varied yet consistent scenes based on identical input.
    }
    \label{fig:diversity}
\end{figure*}

\subsection{Interactive System}
\label{sec:interactive}
We develop an interactive control system (Fig.~\ref{fig:main_method}\textcolor{red}{(a)}) that prioritizes user-directed scene graph manipulation. The core component is a graphical interface where users can precisely construct and modify scene graphs through intuitive operations such as node addition, deletion, and position adjustment. This direct manipulation ensures fine-grained control over the scene generation. Additionally, users can provide text prompts, which are processed by large language models to generate corresponding scene graphs (Fig.~\ref{fig:prompt_generation}). These scene graphs are then used as input to our method to generate the final 3D scenes. Details on the design of system-level prompts can be found in the supplementary materials.

\section{Experimental Results}
\label{sec:experimental_results}

\subsection{Data Preparation}
Due to the lack of paired scene graph data for existing 3D outdoor LiDAR scenes, we generate scene graph data from each scene in the CarlaSC~\cite{carlasc} dataset, creating a new dataset termed \textit{CarlaSG}.
Based on the scene graph formulation discussed in Sec~\ref{sec:sg_formulation}, we extract a 3D scene graph from each 3D semantic map in CarlaSC and project it onto 2D.
Figure~\ref{fig:teaser} provides an example of a 3D outdoor scene alongside its scene graph. 
Notably, as the spatial distribution of sidewalks and ground closely aligns with the road layout, we merge the \textit{Ground} and \textit{Sidewalk} classes in the original CarlaSC dataset with the \textit{Road} class and mark them as \textit{Road}. 
We further categorize the roads into five types: \textit{Straight Road}, \textit{T-Junction}, \textit{Crossroad}, \textit{Bend Road}, and \textit{Others}. Additional details of processing techniques are available in the supplementary materials.

\begin{figure*}[t]
  \centering
  \includegraphics[width = 1\textwidth]
  {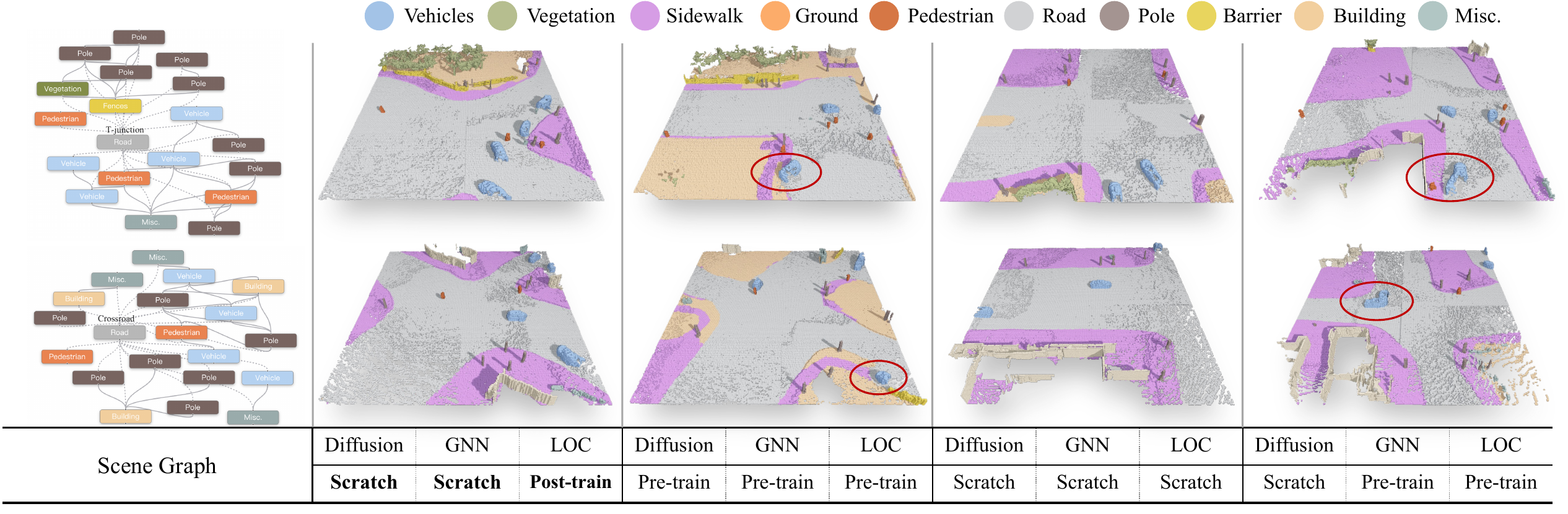}
    \vspace{-2em}
    \caption{\textbf{Impact of Different Training Strategies.} Models trained with the second and last strategies exhibit issues such as vehicles positioned on sidewalks, overlapping objects, and inconsistencies in capturing object quantities. The third strategy generates semantically reasonable scenes but struggles with accurately matching object quantities and road types to the scene graph. In contrast, the first strategy produces high-quality scenes with good alignment to the input scene graph, thus we choose the first strategy to train our networks.}
    \label{fig:different_architecture}
    \vspace{-1em}
\end{figure*}

\subsection{Evaluation Protocols}
We evaluate our approach from two aspects: assessing the quality of generated scenes and measuring the alignment between the generated scenes and their corresponding scene graphs. 
Additionally, we conduct a user study to perceptually 
evaluate the scene graphs' alignment of the generated scenes. 
All experiments are performed on a testing set with \textit{1k} randomly selected scene graphs. Details on the evaluation metrics and examples of the user study can be found in the supplementary materials.

\noindent \textbf{Scene Quality Evaluation.} 
We follow the evaluation protocols from~\cite{pdd} to assess scene quality. We use mean Intersection over Union (mIoU) and mean Accuracy (MA) to evaluate semantic plausibility. Additionally, we measure feature similarity with Fréchet 3D Distance (F3D)~\cite{pdd}, which computes the Fréchet distance between generated and real scenes in a pre-trained 3D CNN-based autoencoder's feature space.

\noindent \textbf{Control Capacity.} 
We evaluate the alignment between generated object counts and scene graph node counts using Mean Absolute Error (MAE) and the Jaccard index. MAE quantifies numerical discrepancies, while the Jaccard index measures the overlap in object types, reflecting how closely the generated scenes match the scene graph structure.

\noindent \textbf{User Study.} We use the Differential Mean Opinion Score (DMOS)~\cite{762345}, a subjective rating method, to evaluate the alignment of generated scenes with input scene graphs, considering object quantity, positioning, and road type.

\subsection{Experimental Settings}
\noindent \textbf{Training and Inference Settings.} 
During joint training of the 2D diffusion model and the GNN, we apply data augmentation and include 10\% unconditional data for the diffusion model, along with a 30\% feature mask on the GNN input to simulate scenarios where some users may not provide positional information for certain nodes. During inference, we set the Gumbel temperature $\tau$ in the allocation module to 2.0 to introduce randomness in generated scenes.
Further details on learning rates, batch size, and other training parameters are in the supplementary material.

\noindent \textbf{Network Architecture.} We employ a diffusion model and GNN in a joint training setup. The 2D/3D diffusion models use 3D-UNet~\cite{3dunet} as the backbone which is often used in outdoor understanding~\cite{li2020campus3d}, while the GNN consists of a two-layer GAT encoder~\cite{gat}. 

\noindent \textbf{Comparison Baselines.} 
As discussed in Sec.~\ref{sec:related_work} and the supplementary material, adapting indoor scene generation methods to outdoor environments is non-trivial due to fundamental differences. Moreover, such adaptations would significantly alter their original pipeline, making direct comparisons less meaningful. Therefore, we consider three baselines: (1) A large language model (LLM)~\cite{ye2024language, zhao2023graphtext} extracts embeddings from the graph's textual description, followed by a 2D deconvolution to align with the downstream 2D diffusion model. Details of the operations are provided in the supplementary materials; (2) Scene Graph to Image (SG2Im)~\cite{johnson2018image}, a GAN-based method for generating images from scene graphs, which we adapt to produce the BEM from scene graphs; and (3) an unconditional generation (Uncon-Gen)~\cite{pdd} model without scene graph conditioning.

\subsection{Main Results}

\textbf{Qualitative Results.} Figure~\ref{fig:main_comparison} shows the 3D outdoor scenes generated separately using our method and baseline methods~\cite{ye2024language, zhao2023graphtext}, based on three different scene graphs.
The results demonstrate that our method effectively captures the object quantities specified in the scene graph and the road type information. In contrast, the scenes generated by the LLM and SG2Im methods show significant discrepancies in object counts across most categories, and the generated road types differ substantially from the intended configurations. 

\noindent \textbf{Quantitative Results.}
Table~\ref{tab:main_results} compares our method with baselines. In Scene Quality, Uncon-Gen, LLM, and our method perform comparably, while SG2Im lags behind. Meanwhile, in Control Capacity, our method outperforms all baselines across metrics, achieving low MAE values below 1.0, demonstrating precise control over object quantities. In contrast, SG2Im has a higher MAE (0.97), and the LLM baseline yields 1.44, over twice our method’s 0.63, indicating a significant accuracy gap. Additionally, our method achieves a higher Jaccard Index, reflecting its effectiveness in capturing object categories from scene graphs across diverse scenes.

\noindent \textbf{Generation Diversity.} To validate that our method produces diverse outputs rather than strictly memorizing scenes 
based on the scene graph, we generate scenes three times using the same scene graph. The results are shown in Figure~\ref{fig:diversity}. The outcomes demonstrate that our method can generate varied scenes even when conditioned on the same scene graph, yet each generated scene remains consistent with the structural and categorical information provided in the scene graph. This confirms that our method introduces randomness in the generation process while maintaining alignment with the input scene graph.

\subsection{Ablation Experiments and User Studies}
\begin{figure}[t]
    \centering
    \includegraphics[width = 1.0\linewidth]
    {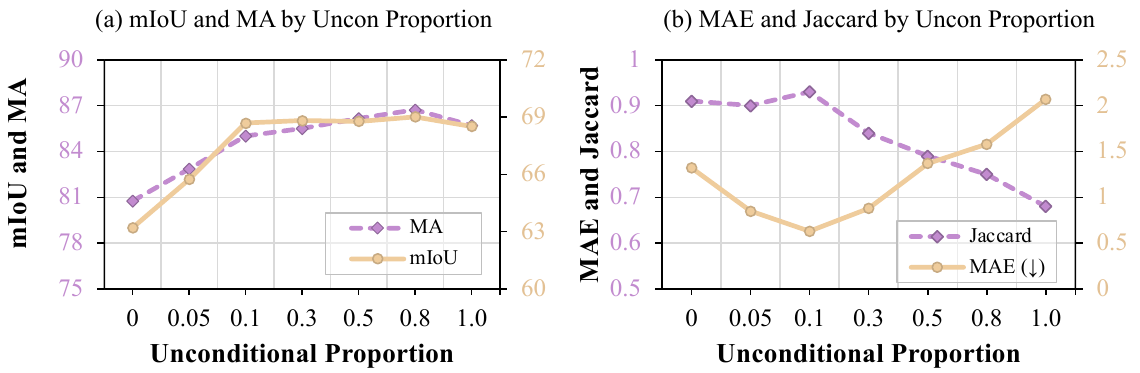}
    \vspace{-2em}
    \caption{\textbf{Unconditional Proportion \textit{v.s.} Generation Quality and Control.} Evaluation mIoU, MA, Jaccard Index, and MAE as the unconditional proportion varies during diffusion training. Considering the trade-off between scene quality and control, we choose 0.1 as the balance point.}
    \label{fig:LineChart}
\end{figure}

\noindent \textbf{Unconditional Proportion.} We examine the effect of the unconditional proportion in diffusion training, as shown in Figure~\ref{fig:LineChart}. Results indicate that scene quality (mIoU and MA) improves as the unconditional proportion increases, with a noticeable bottleneck at 0.1. While further increases lead to marginal improvements in scene quality, they come at the cost of reduced control capacity, as reflected by worsening Jaccard Index and MAE. To balance scene quality and control capacity, we set the unconditional proportion to 0.1 in our experiments.

\begin{figure}[t]
    \centering
    \includegraphics[width = 0.9\linewidth]
    {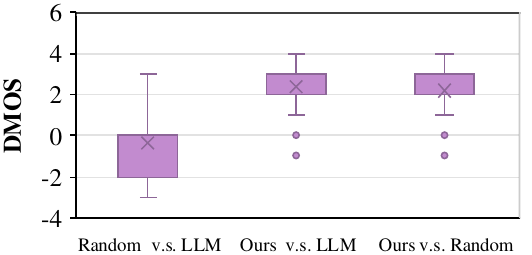}
    \vspace{-0.5em}
    \caption{\textbf{User Study: DMOS Comparison of Scene Generation Methods.} Our method aligns well with scene graph specifications.}
    \label{fig:box_plot}
    \vspace{-1em}
\end{figure}

\noindent \textbf{Effect of Auxiliary Tasks.} We evaluate the impact of adding edge reconstruction and node classification as auxiliary tasks to the GNN during joint training with the diffusion model. 
As shown in Table~\ref{tab:different_tasks}, both tasks yield the best performance, with a low MAE of 0.63 and a high Jaccard Index of 0.93. Removing either task leads to notable drops in performance, particularly in the Jaccard Index. Omitting both results in further declines. This shows that both tasks contribute to improved alignment in scene generation.

\noindent \textbf{Different Training Strategies.} We explore alternative training strategies for our method: (a) pre-train the diffusion model, GNN, and localization head (LOC), then freeze GNN and LOC while fine-tuning the diffusion model; (b) end-to-end training of all components from scratch; (c) pre-train GNN and LOC, freeze their parameters, and train the diffusion model from scratch; and (d) jointly train the diffusion model and GNN from scratch, then freeze GNN and post-train LOC. As shown in Table~\ref{tab:different_stratigies} and Figure~\ref{fig:different_architecture}, strategy (d) achieves the best performance. Strategies (a) and (c) show semantic inconsistencies, while (b) generates scenes of reasonable quality but struggles with object quantity and road type alignment. Joint training of the diffusion model and GNN (d) allows the diffusion model to learn scene structure in sync with encoded features, while post-training LOC assigns precise object positions without disrupting learned structural relationships, achieving a balance between semantic coherence and quantity control.

\noindent \textbf{User Study.} We generate 100 pairs of scenes and conduct user studies with 20 subjects. 
Each user scores paired scenes based on object quantity, positioning, and road type accuracy relative to their scene graphs. The resulting Differential Mean Opinion Score (DMOS), shown in Figure~\ref{fig:box_plot}, indicates that our method outperforms the baselines. 
Additionally, we conduct a one-tailed paired t-test on the MOS score difference among three methods. In this test, the null hypothesis is that our generation method does not possess a higher score than baseline methods.
The results support the rejection of the null hypothesis at a significance level of $p < 10^{-3}$, indicating that our method statistically performs better than both baselines with high confidence.

\begin{table}[t]
\centering
\caption{\textbf{Impact of Auxiliary Tasks on Generation Performance.} Comparison of MAE and Jaccard Index w/ and w/o edge reconstruction and node classification tasks in the GNN. Including both tasks yields the best alignment with the scene graph.}
\vspace{-1em}
\scriptsize 
\setlength{\tabcolsep}{9pt}
\begin{tabular}{cc|cc}
\toprule
Reconstruction            & Classification            & MAE (\(\downarrow)\)           & Jaccard       \\ \midrule \midrule
\ding{51} & \ding{51} & \textbf{0.63} & \textbf{0.93} \\
\ding{51} & \ding{55}     & 0.89          & 0.84          \\
\ding{55}     & \ding{51} & 0.80          & 0.83          \\
\ding{55}     & \ding{55}     & 0.79          & 0.81         \\ \bottomrule
\end{tabular}
\label{tab:different_tasks}
\vspace{5mm}
\end{table}

\begin{table}[t]
\centering
\caption{\textbf{Comparison of Training Strategies for Our Method.} The bolded row is our adopted strategy.}
\vspace{-1em}
\scriptsize 
\setlength{\tabcolsep}{9pt}
\resizebox{0.47\textwidth}{!}{
\begin{tabular}{ccc|cc}
\toprule
Diffusion        & GNN          & LOC             & MAE (\(\downarrow)\) & Jaccard       \\ \midrule \midrule
Pre-train        & Pre-train        & Pre-train           & 0.79                 & 0.88          \\
Scratch          & Scratch          & Scratch             & 1.01                 & 0.82          \\
Scratch          & Pre-train        & Pre-train           & 0.95                 & 0.90          \\
\textbf{Scratch} & \textbf{Scratch} & \textbf{Post-train} & \textbf{0.63}        & \textbf{0.93} \\ \bottomrule
\end{tabular}}
\label{tab:different_stratigies}
\end{table}

\section{Conclusion}
In this work, we propose a solution that integrates an interactive system, BEV Embedding Map, and diffusion generation to enable controllable 3D outdoor scene generation. The challenges stem from complex outdoor landscapes with rich information and structural diversity. Our approach utilizes scene graphs to transition information from sparse to dense representations. Coupled with the interactive system, it enables users to intuitively and concisely generate their desired 3D outdoor scenes. Comparative experiments demonstrate that our method achieves more accurate object quantities and alignment with the input scene graph. These results indicate that our approach is a robust and effective solution for controllable 3D outdoor scene generation.
\label{sec:conclusion}

{
    \small
    \bibliographystyle{ieeenat_fullname}
    \bibliography{main}
}

\end{document}